\documentclass[submission,copyright,creativecommons]{eptcs}
\providecommand{\event}{FMAS 2024} 

\usepackage[utf8]{inputenc}
\usepackage[T1]{fontenc}
\usepackage[english]{babel}
\usepackage{multicol}
\usepackage{url}
\usepackage{float}

\usepackage{tikz,xcolor}
\usepackage{tkz-graph}
\usetikzlibrary{fit,arrows.meta,decorations,shapes.geometric,decorations.pathmorphing}
\usetikzlibrary{shapes,arrows.meta, automata}
\usetikzlibrary{positioning}

\usepackage{color}
\usepackage{amsfonts}
\usepackage{amsmath}
\usepackage{amssymb}
\usepackage{mathtools}
\usepackage{graphicx}
\usepackage{wrapfig}
\usepackage{caption}
\usepackage{subcaption}
\usepackage{pstricks}

\usepackage[textsize=small]{todonotes}

\title{Cross-Layer Formal Verification of Robotic Systems\footnote{This study has received financial support from the French government in the framework of the France 2030 programme IdEx université de Bordeaux / RRI ROBSYS}}
\author{Sylvain Raïs$^{1,2}$, Julien Brunel$^1$, David Doose$^1$ and Frédéric Herbreteau$^2$
  \institute{$^1$ ONERA DTIS, Université de Toulouse, France}
  \institute{$^2$ Univ. Bordeaux, CNRS, Bordeaux INP, LaBRI, UMR 5800, 33400, Talence, France}
  \email{$^1$ firstname.lastname@onera.fr, $^2$ firstname.lastname@u-bordeaux.fr}
}

\def\titlerunning{Cross-layer Formal Verification of Robotic Systems}
\def\authorrunning{S. Raïs, J. Brunel, D. Doose \& F. Herbreteau}

\newcommand{\tsname}[1]{\ensuremath{\mathsf{#1}}}

\ifpdf
  \usepackage{underscore}         
  \usepackage[T1]{fontenc}        
\else
  \usepackage{breakurl}           
\fi

\usepackage{listings}
\definecolor{gray}{HTML}{666666}		
\definecolor{lightblue}{HTML}{006699}		
\definecolor{darkblue}{HTML}{003355}		
\definecolor{lightgreen}{HTML}{669900}		
\definecolor{bluegreen}{HTML}{33997e}		
\definecolor{orange}{HTML}{e2661a}		
\definecolor{purple}{HTML}{7d4793}		
\definecolor{commentscolor}{rgb}{0,0.6,0}
\definecolor{numbercolor}{rgb}{0.5,0.5,0.5}
\definecolor{stringcolor}{rgb}{0.2,0.6,0.8}
\definecolor{backcolor}{rgb}{0.93,0.93,0.93}
\definecolor{keywordcolor}{rgb}{0,0,0.75}
\definecolor{keywordcolorbis}{rgb}{0,0.6,0.5}
\definecolor{identifiercolor}{rgb}{0.25,0.25,0.25}
\definecolor{violet}{rgb}{0.6,0.4,0.8}

\lstdefinelanguage{rotboLanguage}{
    morekeywords = [1]{skillset},
    morekeywords = [2]{data, resource, event, skill},
    morekeywords = [3]{state, transition, initial, guard, input, output, precondition, start, invariant, progress, update, interrupt, success, failure, effect},
    morekeywords = [4]{period},
    otherkeywords = {!,>,<,.,;,-,=,not,and,true,false,0,1,2,3,4,5,6,7,8,9},
    morekeywords = [5]{!,-,>,<,=,.,0,1,2,3,4,5,6,7,8,9},
    morekeywords = [6]{not,and},
    morekeywords = [7]{true,false},
    morecomment = [l]{//},
    morecomment = [s]{/*}{*/},
    morecomment = [s]{/**}{*/},
    keywordstyle = [1]\color{lightblue}\bfseries,
    keywordstyle = [2]\color{lightblue}\bfseries,
    keywordstyle = [3]\color{lightblue}\bfseries,
    keywordstyle = [4]\color{lightblue}\ttfamily,
    keywordstyle = [5]\color{orange}\ttfamily,
    keywordstyle = [6]\color{lightblue}\ttfamily,
    keywordstyle = [7]\color{darkblue}\ttfamily,
    commentstyle = \color{lightgreen}
}

\lstdefinestyle{rotboLanguage}{
    language=rotboLanguage,
    basicstyle=\ttfamily\small\color{black},
    identifierstyle=\color{identifiercolor},
    commentstyle=\color{commentscolor},
    numberstyle=\scriptsize\color{numbercolor},
    stringstyle=\color{stringcolor},
    keywordstyle=\bfseries\color{keywordcolor},
    keywordstyle={[2]\bfseries\color{keywordcolorbis}},
    breakatwhitespace=true,
    breaklines=true,
    keepspaces=true,
    numbers=left,
    numbersep=0pt,
    showspaces=false,
    showstringspaces=false,
    showtabs=false,
    tabsize=2,
}

\begin{document}
\maketitle

\begin{abstract}
  Robotic systems are widely used to interact with humans or to perform critical tasks. As a result, it is imperative to provide guarantees about their behavior. Due to the modularity and complexity of robotic systems, their design and verification are often divided into several layers. However, some system properties can only be investigated by considering multiple layers simultaneously.
  We propose a cross-layer verification method to verify the expected properties of concrete robotic systems. Our method verifies one layer using abstractions of other layers. We propose two approaches: refining the models of the abstract layers and refining the property under verification. A combination of these two approaches seems to be the most promising to ensure model genericity and to  avoid the state-space explosion problem.
\end{abstract}

\section{Introduction}
\label{sec:introduction}

The design and development of modern robotic systems is a complex issue, as it brings together many fields of research. Moreover, these robotic systems are intended to interact with humans or to be deployed in critical sites. Therefore, it is essential to provide guarantees for the operation of these systems.
Formal methods are widely used to assert the reliability of critical systems.
They provide strong proof-based guarantees that the verified system behaves accordingly to the specifications.
In the context of robotic systems, several modeling tools and formalisms have been developed to verify properties, either online~\cite{ROSRV} or offline~\cite{SYSWIDE,PATT}.

On the other hand, in order to improve the design of robotic systems, state-of-the-art approaches rely on multi-layer architectures as they provide powerful abstraction to develop each layer independently of the others. Such a design facilitates the development of robotic systems, improves their modularity and enables each layer to be (formally) verified separately. These advantages help to implement complex behaviors such as fault tolerance~\cite{LeitePM18} and facilitate the reuse of robotic system code. Note that several multi-layer design standards exist within the robotics research community: five-layer pyramid design~\cite{ALCACER2019899}, four-layer design~\cite{SkiROS}, three-layer pyramid design~\cite{PEDERSEN2016282, SCHOU201872}, and more.
Among these classical designs, the three-layer architecture shown in Figure~\ref{strata} is a promising and widely used approach because it provides a modular design while minimizing the number of layers. In this architecture, the decision layer deals with the robot's decision-making and planning processes (e.g. a user interface or a "smart" program). The executive layer provides an abstract interface to the functional layer via the concept of skills~\cite{articleC, SkiROS, SCHOU201872, PEDERSEN2016282}. And the functional layer corresponds to low-level task processing.

\noindent
\begin{minipage}{.62\textwidth}
    {\tiny\begin{lstlisting}[language=rotboLanguage, caption=An example of RobotLanguage design, label=case_study]
skillset custom_robot {
    resource {
        motion { state { On Off } initial Off transition all }
        battery { state { Normal Critical } initial Normal transition all }
    }
    skill goto {
        input {distance: Integer}
        output position: Position
        precondition { (motion == Off) && (battery != Critical) }
        start motion -> On
        invariant {in_movement { guard motion == On }}
        interrupt {effect {motion -> Off}}
        success {arrived {effect {motion -> Off}}}
        failure {blocked {effect {motion -> Off}}}
    }
}
\end{lstlisting}}
\end{minipage}
\hfill
\begin{minipage}{.35\textwidth}
    \begin{figure}[H]
        \centering
        \includegraphics[scale=0.08]{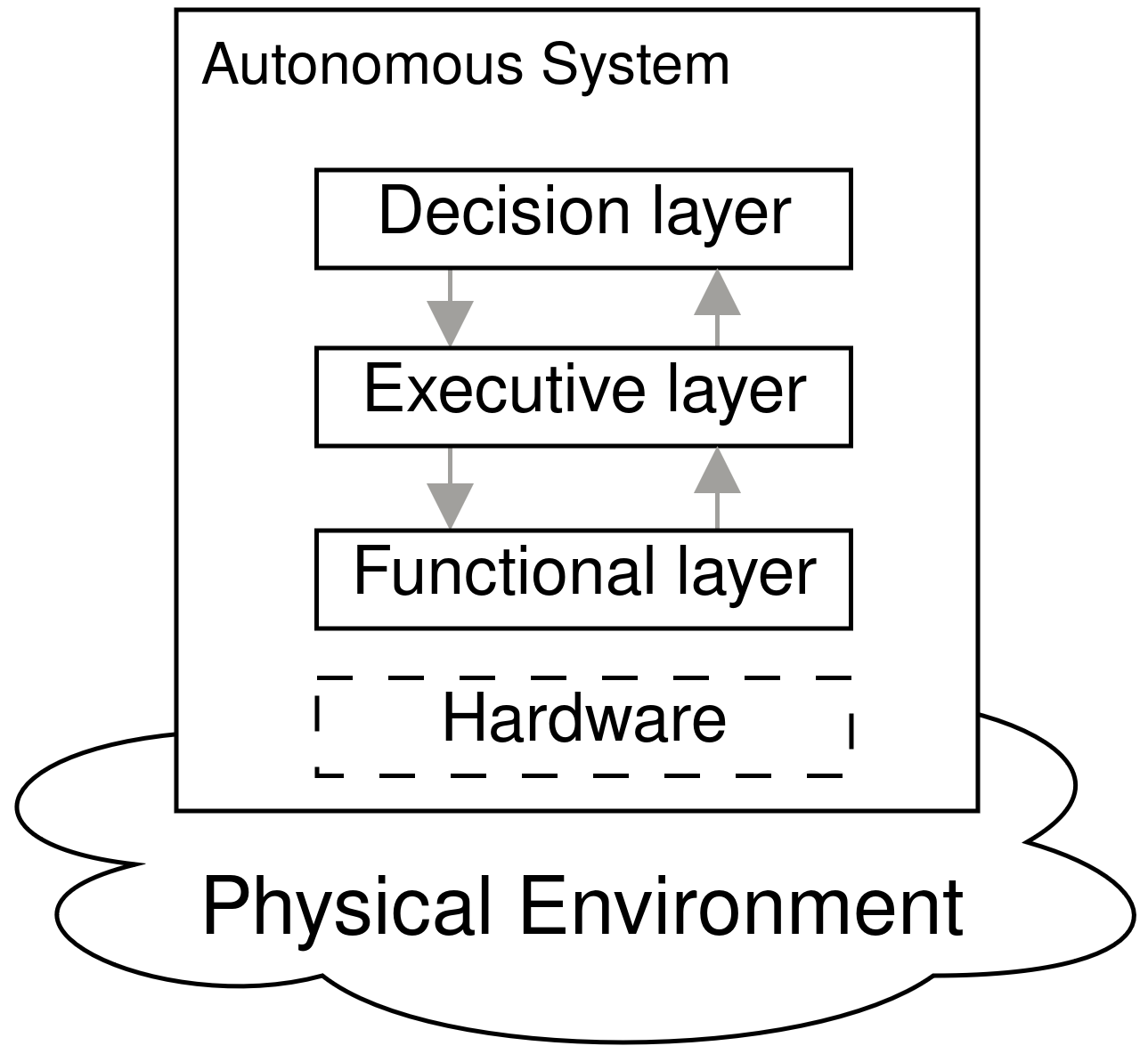}
        \caption{Three-layer architecture}\label{strata}
    \end{figure}
\end{minipage}

\medskip

In practice, it is impossible to verify the whole system at once, due to the complexity of robotic systems, or the incompatibility of certain theories that make verification undecidable.
Specific formalisms and techniques have been developed for the design and the verification of each layer separately. However, the compartmentalization of the different analyses is an obstacle to complete system analysis because these formalisms cannot always be combined. In fact, some of the operating characteristics of the system must consider multiple layers in order to be studied.

The present work, which is part of a Ph.D. thesis, aims to provide an offline cross-layer verification method based on the three-layer design in Figure~\ref{strata}.
Our method uses RobotLanguage \footnote{\url{https://onera-robot-skills.gitlab.io/index.html}} \cite{articleA,articleB,articleC}, an interesting framework for designing reliable robotic systems. RobotLanguage provides a formal language to model the executive layer, a formal offline verification of predefined properties on this model, and an automatic code generation from the model to implement this layer. 
Our approach is based on a RobotLanguage model of the executive layer, and extends it with abstract models of the other layers in order to verify properties of the whole system.
In general, abstract models are not refined enough to verify robotic systems.
We introduce two complementary approaches: one consists in refining the model, and the other consists in refining the property.
We illustrate the relevance of our techniques on an example.
Our method is not specific to the three-layer architecture in Figure~\ref{strata}, and can be used with other multi-layer designs.

\section{Related Works}
\label{sec:related}

Several formal frameworks have been defined to support the design and the verification of robotic systems, such as RobotLanguage. PROSKILL \cite{Proskill} gathers the specifications of the decisional layer, the executive layer and a part of the functional layer (see Figure~\ref{strata}), and allows to verify temporal and timed properties both offline and online. However, PROSKILL provides a monolithic design for robotic systems and does not benefit from the advantages of a multi-layer design. Our method is based on the multi-layer design, preserving the modularity gained by this design, and thus fits well to our real robotic systems.

On the other hand, RobotLanguage comes with a tool, SkiNet \cite{Skinet}, which provides a translation to Petri nets to perform offline formal verification of temporal properties. This tool has also been extended \cite{PerdRunTime} to verify temporal properties online in order to address the state-explosion problem. However, SkiNet only verifies properties of the executive layer only, while our work aims to provide a multi-layer verification method.

\section{Cross-Layer Verification}
\label{sec:problem-overview}
RobotLanguage has been developed to design the executive layer of robotic systems. After a brief introduction, we describe the formalism used to model these systems. Next, we explain how to model each system layer and how to incorporate all models for formal verification. Finally, we present a method for systematically verifying multi-layer systems, illustrated with an example.

\subsection{Introduction to RobotLanguage}

Modern approaches to formal robotic system design are based on skills and resources\cite{Proskill, articleA, SkiROS, PEDERSEN2016282}. Skills are basic actions provided by the executive layer to implement complex behaviors in the decision layer. For example, Listing\ref{case_study} defines one skill: \texttt{goto}, which moves a robot a given distance. Resources represent physical features used by skills, such as \texttt{motion} and \texttt{battery} in Listing~\ref{case_study}. The resource \texttt{battery} tracks levels, while \texttt{motion} monitors movement. In RobotLanguage, each group of skills and their shared resources forms a \emph{skill set}, such as \texttt{custom_robot} in Listing~\ref{case_study}.

The skill \texttt{goto} is an abstraction of the actual code executed at the functional layer. 
In RobotLanguage the system designer specifies conditions for starting a skill (\texttt{precondition}), conditions that should remain true during execution (\texttt{invariant}), and resource updates (\texttt{start}, \texttt{effect}).

RobotLanguage includes a toolset\footnote{\url{https://onera-robot-skills.gitlab.io/}} that translates models into executable C++ code using the ROS2 middleware. This code creates one ROS2 node per skill set and several topics to manage communication between the executive and decision layers. In addition, the generated code verifies conditions (\texttt{precondition}, \texttt{invariant}) and applies effects (\texttt{start}, \texttt{effect}) specified in RobotLanguage. The programmer is responsible for implementing the functional layer in specific hook functions, whose prototypes are generated from the RobotLanguage design.

\subsection{Modeling Formalism}\label{for_mod}

In this paper, a model consists of a finite set $M = \{S_1, \dots, S_k\}$ of finite labeled transition systems. Each transition system $S_i = (Q_i, q_i^0, \Sigma_i, T_i)$ consists of a finite set of states $Q_i$, a distinguished initial state $q_i^0$, a finite alphabet of events $\Sigma_i$ and a transition relation $T_i \subseteq Q_i \times \Sigma_i \times Q_i$ where edges are labeled by events from $\Sigma_i$.
Note that the transition systems may have common events on which they synchronize. Let $\Sigma = \bigcup_{i \in [1;k]} \Sigma_i$.
A global state of $M$ is a tuple $(q_1, \dots, q_k)$ of states, one for each transition system in $M$. 
The initial global state is $(q_1^0, \dots, q_k^0)$. There exists a global transition $(q_1, \dots, q_k) \xrightarrow{a} (q_1', \dots, q_k)$ with $a \in \Sigma$ if for each $S_i$ such that $a \in \Sigma_i$, there exists a transition $(q_i, a, q_i') \in T_i$, and $q_i' = q_i$ for every $S_i$ such that $a \notin \Sigma_i$. A global run is a sequence of global transitions starting from the initial global state.

As an example, consider the model consisting of two transition systems: $S$ in Figure~\ref{skill_state_machine} and $F$ in Figure~\ref{fig:goto_default_func}. These two transition systems synchronize on their common labels. Thus, any run in this model consists of asynchronous solid and zigzag transitions from $S$, or dotted and dashed transitions that synchronize $S$ and $F$.

\subsection{Executive Layer Modeling}

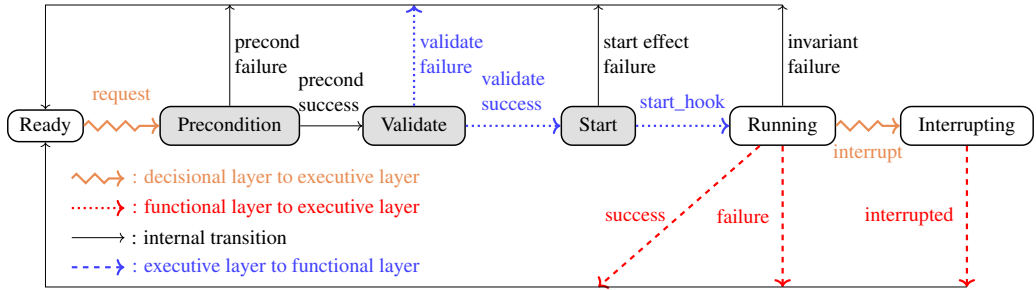
\begin{figure}[t]
    \center
    \scalebox{0.7}{\begin{tikzpicture}
    \tikzstyle{call}=[rectangle, rounded corners=6pt, thick, draw]
    \tikzstyle{state}=[draw, rounded corners=6pt, thick, fill=gray!20]
    \tikzstyle{fail}=[draw, rectangle, draw=red, thick, fill=red!20]
    \tikzstyle{func}=[draw, rounded corners=6pt, draw=blue, thick, fill=blue!20]
    \tikzstyle{success}=[draw, rectangle, draw=green, thick, fill=green!20]
    \tikzstyle{failure}=[draw, rectangle, draw=red, thick, fill=red!20]
    \tikzstyle{hook}=[very thick, blue!75]
    \tikzstyle{functional}=[very thick, red]
    \tikzstyle{event}=[decorate, decoration={zigzag, post length=0.25cm}, very thick, color=orange!75]

    \def\dx{3.5cm}
    \def\dy{1.75cm}
    \def\h{1.75}

    \node[call] (C) at (0,0) {~Ready ~};
    \node[state] (P) at (\dx,0) {\begin{tabular}{c}Precondition\end{tabular}};
    \node[state] (V) at (2*\dx,0) {\begin{tabular}{c}Validate\end{tabular}};
    \node[state] (S) at (3*\dx,0) {\begin{tabular}{c}Start\end{tabular}};
    \node[call] (R) at (4*\dx,0) {\begin{tabular}{c}Running\end{tabular}};
    \node[call] (I) at (5*\dx,0) {\begin{tabular}{c}Interrupting\end{tabular}};

    \draw[->, event] (C) -- node[above=0.25cm] {request} (P);
    \draw[->] (P) -- 
        node[above=0cm] {
        \begin{tabular}{l}
            precond \\
            success
        \end{tabular}
    } (V);
    \draw[->, hook, dotted] (V) -- node[above] 
        {\begin{tabular}{l} validate\\success \end{tabular}} (S);
    \draw[->, hook, dotted] (S) -- node[above=0.15cm] {start\_hook} (R);
    \draw[->, hook, dotted] (V) -- node[right=-0.25cm] 
        {\begin{tabular}{l} validate\\failure \end{tabular}} (2*\dx, \h * \dy * 0.75);
    \draw[->] (R) -- node[right=-0.25cm] {
        \begin{tabular}{l}
            invariant \\
            failure
        \end{tabular}
    }  (4*\dx, \h * \dy * 0.75);
    \draw[->, event] (R) -- node[below=0.2cm] {interrupt}(I);
    \draw[->, functional, dashed] (R) -- node[left=0.1cm] {success} (3*\dx, -\h * \dy);
    \draw[->, functional, dashed] (R) -- node[left=0.1cm] {failure} (4*\dx, -\h * \dy);
    \draw[->, functional, dashed] (I) -- node[left=0.1cm] {interrupted} (5*\dx, -\h * \dy);

    \draw[->] (P) -- node[right=-0.25cm] {
        \begin{tabular}{l}
            precond \\
            failure
        \end{tabular}
    } (\dx, \h * \dy * 0.75);
    \draw[->] (S) -- node[right=-0.25cm] {
        \begin{tabular}{l}
            start effect \\
            failure
        \end{tabular}
    } (3*\dx, \h * \dy * 0.75);
    \draw[->] (4*\dx, \h * \dy * 0.75) -- (0, \h * \dy * 0.75) -- (C);

    \draw[->] (5*\dx, -\h * \dy) -- (0, -\h * \dy) -- (C);

    \def\lx{0.5cm}
    \def\ly{-1.5cm}
    \def\dl{0.75cm}

    \draw[->, event] (\lx, \ly + 0.5cm) -- node[right=0.5cm] {: decisional layer to executive layer} (\lx + 1cm, \ly + 0.5cm);
    \draw[->, functional, dotted] (\lx, \ly-\dl*0.75+0.5cm) -- node[right=0.5cm] {: functional layer to executive layer} (\lx + 1cm, \ly-\dl*0.75+0.5cm);
    \draw[->] (\lx, \ly-2*\dl*0.75+0.5cm) -- node[right=0.5cm] {: internal transition} (\lx + 1cm, \ly-2*\dl*0.75+0.5cm);
    \draw[->, hook, dashed] (\lx, \ly-3*\dl*0.75+0.5cm) -- node[right=0.5cm] {: executive layer to functional layer} (\lx + 1cm, \ly-3*\dl*0.75+0.5cm);

\end{tikzpicture}}
    \caption{Control flow graph of a skill}
    \label{skill_state_machine}
\end{figure}

First, we explain how to model the executive layer by describing the execution of a skill through the transition system in Figure~\ref{skill_state_machine}.
During its execution, the skill transitions through several states, depending on internal actions (plain transitions), or on interactions with the decision layer (zigzag transitions) or with the functional layer (dashed/dotted transitions). 
The execution begins in the state \texttt{Ready}, and spans into three phases.
First, on reception of \texttt{request} from the decision layer, the state of the system is checked at the executive layer (\texttt{precond}) and the functional layer (\texttt{validate}).
If these conditions are satisfied, \texttt{start_hook} triggers the execution of the functional layer, switching the state to \texttt{Running}.
Finally, the execution can terminate in a success or a failure, triggered by the functional layer, or it can be interrupted by the decision layer. In each case, the functional layer notifies the executive layer by calling \texttt{success}, \texttt{failure}, or \texttt{interrupted}. 
The execution can also stop if an invariant is violated. These invariants are monitored by the code that is automatically generated from the RobotLanguage design. 
We refer the reader to~\cite{articleC} for more details on the semantics of invariants in RobotLanguage which is beyond the scope of this paper.

For a given skill set like in Listing~\ref{case_study}, a model using instances of the transition system in Figure~\ref{skill_state_machine} for each skill can formally verify some properties at the executive layer, such as “skill \texttt{goto} can be executed.” However, this model is not refined enough to verify more specific properties, such as “skill \texttt{goto} cannot be executed infinitely often”, which is expected to hold, since our RobotLanguage design in Listing~\ref{case_study} lacks a skill to recharge the battery.

\subsection{Multi-Layer Modeling}

We aim to extend the model in Figure~\ref{skill_state_machine} (called $S$ in the sequel) with models of the functional and decision layers. 
For now on, we will concentrate on the functional layer, as the approach that we present hereafter straightforwardly applies to the decision layer.

From Section~\ref{for_mod}, it comes that a model of the functional layer should conform to a \emph{synchronization interface} that will enable communication between the model of the functional layer, and the model of the executive layer, through event synchronizations.
More specifically, a model of the functional layer should synchronize on events 
\textcolor{blue}{validate success}, \textcolor{blue}{validate failure}, \textcolor{blue}{start hook}, \textcolor{red}{success}, \textcolor{red}{failure} and \textcolor{red}{interrupted} with the transition in Figure~\ref{skill_state_machine}.

\begin{figure}[t]
    \centering
    \begin{subfigure}[b]{0.35\textwidth}
        \centering
        \scalebox{1}{\scalebox{0.8}{
    \begin{tikzpicture}[scale=0.9,->,>=stealth',shorten >=1pt,auto,node distance=2.5cm,initial text={},
    every initial by arrow/.style={->}]
    {\fontsize{10}{10}\selectfont
    \tikzstyle{every node}=[scale=1,inner sep=1pt, outer sep=1pt,line width=1pt]
    \tikzstyle{every edge}=[scale=1,line width=1pt, draw]
    \tikzstyle{every state}=[draw,text=black]
    \node[state, initial,rectangle]  (Idle)  at (0,0)                  {$\tsname{Idle}$};

    \draw [blue, dotted] (Idle.north west) |- (-1.5,1)
        node [above=10pt] {}
        node [above=0pt] {\color{blue!75} validate success}
        |- (Idle.north west);
    \draw [blue, dotted] (Idle.north east) |- (1.5,1)
        node [above=10pt] {}
        node [above=0pt] {\color{blue!75} validate failure}
        |- (Idle.north east);        
    
    \draw [red, dashed] (Idle.south west) |- (-1.5,-1)
        node [left=0pt] {\color{red} success}
        |- (Idle.south west);
    \draw [red, dashed] (Idle.south east) |- (1.5,-1)
        node [right=0pt] {\color{red} interrupted}
    |- (Idle.south east);
    \draw [red, dashed] ([shift={(0,-0.25)}] Idle.east) -| (1.5,0)
        node [right=0pt] {\color{red} failure}
    |- ([shift={(0,0.25)}] Idle.east);
    \draw [blue, dotted] ([shift={(0,-0.25)}] Idle.west) -| (-1.5,0)
        node [left=0pt] {\color{blue!75} start\_hook}
    |- ([shift={(0,0.25)}] Idle.west);

    \node (Empty) at (0, -2.5) {};

    }
    \end{tikzpicture}
}}
        \caption{Default functional layer model}
        \label{fig:goto_default_func}
    \end{subfigure}
    \hfill
    \begin{subfigure}[b]{0.6\textwidth}
        \centering
        \scalebox{1}{\scalebox{0.8}{
    \begin{tikzpicture}[scale=0.9,->,>=stealth',shorten >=1pt,auto,node distance=2.5cm,initial text={},
    every initial by arrow/.style={->}]
    {\fontsize{10}{10}\selectfont
    \tikzstyle{every node}=[scale=1,inner sep=1pt, outer sep=1pt,line width=1pt]
    \tikzstyle{every edge}=[scale=1,line width=1pt, draw]
    \tikzstyle{every state}=[draw,text=black]
    \node[state, initial,rectangle]  (Idle)  at (0,0)                  {$\tsname{Idle}$};

    \draw [blue, dotted] (Idle.north west) |- (-2.2,1)
        node [above=10pt] {\black $(blevel \ge 2.0) \land (d > 0)$}
        node [above=0pt] {\color{blue!75} validate success}
        |- (Idle.north west);
    \draw [blue, dotted] (Idle.north east) |- (2.2,1)
        node [above=10pt] {\black $(blevel < 2.0) \lor (d \le 0)$}
        node [above=0pt] {\color{blue!75} validate failure}
        |- (Idle.north east);
    
    \draw [red, dashed] (Idle.south west) |- (-1.5,-1)
        node [shift={(0,0.4)}] [left=0pt] {\color{red} success}
        node [shift={(0,0)}] [left=0pt] {\black $(d \leq 0)$}
        |- (Idle.south west);
    \draw ([shift={(-0.25,0)}] Idle.south) |- (0,-1.25)
        node [below=2pt] {move}
        node [below=10pt] {$(blevel \ge 2.0) \land (d > 0)$}
        node [below=20pt] {\color{black!75} $d := d-1$, $blevel := blevel-2.0$}
        -| ([shift={(0.25,0)}] Idle.south);
    \draw [red, dashed] (Idle.south east) |- (1.5,-1)
        node [right=0pt] {\color{red} interrupted}
    |- (Idle.south east);
    \draw [red, dashed] ([shift={(0,-0.25)}] Idle.east) -| (2.2,0)
        node [shift={(0,0.2)}] [right=0pt] {\color{red} failure}
        node [shift={(0,-0.2)}] [right=0pt] {\black $(blevel < 2.0) \land (d > 0)$}
    |- ([shift={(0,0.25)}] Idle.east);
    \draw [blue, dotted] ([shift={(0,-0.25)}] Idle.west) -| (-2.2,0)
        node [left=0pt] {\color{blue!75} start\_hook}
    |- ([shift={(0,0.25)}] Idle.west);
    }
    \end{tikzpicture}
}}
        \caption{Concrete functional layer model for skill \texttt{goto}}
        \label{fig:goto_func}
    \end{subfigure}
    \caption{Transition systems modeling the skill \texttt{goto} at the functional layer}
    \label{fig:goto_state_machines}
\end{figure}
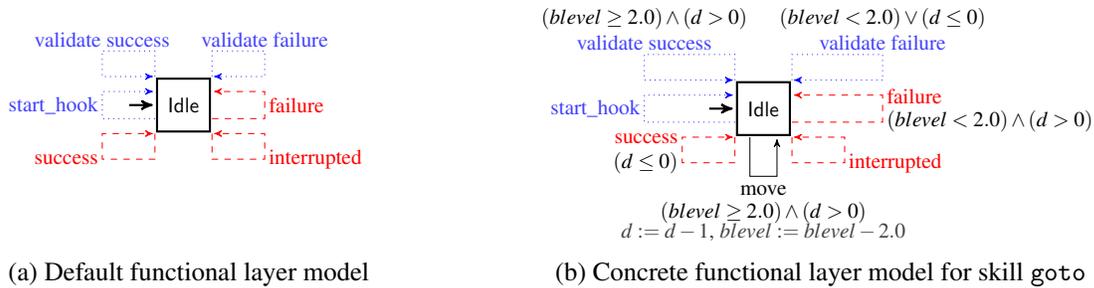

The transition system $F$ in Figure~\ref{fig:goto_default_func} shows a very abstract model of a functional layer that conforms to this synchronization interface.
Note that $F$ allows any sequence of the above mentioned events since
its transitions can be crossed unconditionally.
Hence, any sequence of events that is possible in $S$ is also possible in the model $\{S, F\}$ where $S$ and $F$ synchronize on common labels.
$F$ can be seen as the generic most abstract functional layer model.

Figure~\ref{fig:goto_func} shows another model of the skill \texttt{goto} at the functional layer. 
Observe that this model also conforms to the synchronization interface. 
It further has an internal action \texttt{move} that does not synchronize with $S$: it is asynchronous. 
This model is described as a control graph with two variables: $d$ which is the distance to travel, and $blevel$ which is the battery level (both variables have finite domains).
Note that the variables used at the functional layer partly model the robot's state, while the RobotLanguage resources used in the executive layer (Listing~\ref{case_study}) are abstract knowledge of the robot's state, updated by monitoring the robot.
The model for updating the \texttt{battery} resource according to the actual value of $blevel$ is not shown for the sake of simplicity.
Following our settings described in Section~\ref{for_mod}, we consider the transition systems $F'$ that defines the semantics of control graph in Figure~\ref{fig:goto_func}.
Its states are pairs $(d, blevel)$ of values of the two variables, and transitions $(d, blevel) \xrightarrow{a} (d', blevel')$ take into account the guards and updates on the variables.
Now, observe that due to variables $d$ and $blevel$, the model $F'$ restricts the sequences of events that can occur in a run.
For instance, \textcolor{blue}{validate success} is not possible if the battery level is less than $2.0$.
$F'$ can thus be seen as a refinement of $F$.
As a result, some runs that exist in $S$ do not exist any more in the model $\{S, F'\}$ that synchronizes $S$ and $F'$.

Similarly, we can model the decision layer, with a synchronization interface that is defined by the events \textcolor{orange}{request} and \textcolor{orange}{interrupt}.
We thus obtain a multi-layer model, that consists in transition systems for each skill (as in Figure~\ref{skill_state_machine}) at the executive layer, for each skill at the functional layer (as in Figure~\ref{fig:goto_default_func} or~\ref{fig:goto_func}), as well as transition systems for each resource (as defined in Listing~\ref{case_study}) and a transition system modeling the decision layer.

\subsection{A Method for Cross-layer Verification}

We aim at verifying that “skill \texttt{goto} cannot be executed infinitely often” taking into account a model of the functional layer.
This property can be expressed in LTL as:
\begin{equation}\label{eq:prop}
    \mathsf{F}\, \mathsf{G} \,\, \text{not} \ \mathtt{Running}
\end{equation}
This formula specifies that after some finite amount of time, the robot will never be running.
It does not hold when the functional layer is modeled as in Figure~\ref{fig:goto_default_func}.
We present two approaches for verifying such specifications requiring a multi-layer model.

A first approach consists in considering the refined model of the functional layer in Figure~\ref{fig:goto_func}, where $d$ represents the distance to travel, and $blevel$ tracks the battery level.
In Figure~\ref{fig:goto_func}, the black loop moves the robot one meter ahead, consuming two battery units at the same time. At some point, either the distance $d$ reaches $0$ which leads to a \texttt{success}, or the battery level gets below $2.0$ which leads to a \texttt{failure}. 
Observe that the battery level $blevel$ is set at the initialization of the model. Hence, the battery level eventually becomes insufficient to execute skill \texttt{goto}: it only allows ``\texttt{validate failure}'' and ``\texttt{failure}'' transitions.
As a result, property~\eqref{eq:prop}, that is ``skill \texttt{goto} cannot execute infinitely often'', holds on the refined model.
    
A second approach consists in refining the specification. 
In this approach, we aim at verifying our property: ``skill \texttt{goto} cannot execute infinitely often'' on the model including the abstract representation of the functional layer from Figure~\ref{fig:goto_default_func}, but \emph{with some extra assumptions}.
Coming back to our example, since our RobotLanguage design in Listing~\ref{case_study} does not include a skill to recharge the battery, we can expect the resource \texttt{battery} to be in state \texttt{Critical} after some finite amount of time.
Hence, we can verify that ``skill \texttt{goto} cannot execute infinitely often'' \emph{under the assumption} ``eventually the \texttt{battery} is forever in state \texttt{Critical}''.
This approach consists in refining the LTL formula in~\eqref{eq:prop} to verify the property only on runs which satisfy this assumption.
This is formalized in~\eqref{eq:assumpt}, where \texttt{Critical} corresponds to the state of the resource \texttt{battery} in Listing~\ref{case_study}. 
This formula ensures that if the \texttt{battery} eventually stays in state \texttt{Critical} forever, then, the skill \texttt{goto} is not executed infinitely often.

\begin{equation}\label{eq:assumpt}
    \mathsf{F} \, \mathsf{G} \,\, \texttt{Critical} \implies 
    \mathsf{F}\, \mathsf{G} \,\, \text{not} \ \mathtt{Running}
\end{equation}

Observe that due to the precondition in Listing~\ref {case_study} the transition labeled ``\texttt{precond success}'' in Figure~\ref{skill_state_machine} can only be taken a finite number of times on any run such that the \texttt{battery} eventually stays in state \texttt{Critical} forever. As a result the property in~\eqref{eq:assumpt} holds on the abstract model of the system with the functional layer modeled by the transition system in Figure~\ref{fig:goto_default_func}. Observe that this model does not need any extra variable and is thus much smaller than the model obtained with the first approach.

To validate our approaches, we have translated the transition systems and specifications corresponding to the two approaches, as formulas for the \textit{Tatam} model-checker\footnote{Tatam git repository: \url{https://github.com/DavidD12/tatam}}. The RobotLanguage design in Listing~\ref{case_study} as well as the Tatam models underlying the two approaches above are available on a \href{https://gitlab.com/sylvain.rais24/fmas_2024_s_rais_models}{public repository}\footnote{\url{https://gitlab.com/sylvain.rais24/fmas_2024_s_rais_models}}.
As expected, we have first observed that the property ``skill \texttt{goto} cannot execute infinitely often'' does not hold on the abstract model of the functional layer in Figure~\ref{fig:goto_default_func} as the discharge of the battery is not taken into account. On the other hand, the two approaches above allow to prove that the property holds, either by providing a refined model of the functional layer, or by refining the specification.

We see these two approaches as complementary tools for cross-layer verification of robotic systems.
Refining the property keeps the model small and simple. 
It also yields a simpler counter-example when a property is not satisfied.
However, some properties require a more precise knowledge of the state of the robot.
Then, the first approach should be used to refine (parts of) the model with as few details as possible in order to be able to verify the property under consideration.

\section{Conclusion}
\label{sec:conclusion}

This paper presents a method for cross-layer verification of robotic systems. Our approach consists in verifying one layer using abstractions of the others. We have proposed two approaches to prove a property. One consists in refining the models of the abstract layers, the other consists in refining the property. In practice, the combination of the two approaches seems to be the most promising since it allows to consider as few implementation details as possible in the model, while mitigating the state-space explosion problem.

As future work, we plan to implement our approach in a tool to formally verify RobotLanguage designs using a precise model of the executive layer and abstract models of the decision and functional layers. To obtain a full guarantee approach, we plan to extend our technique to prove that these abstract models correspond to the implementation of the corresponding layers.

\nocite{*}
\bibliographystyle{eptcs}
\bibliography{generic}

\end{document}